\documentclass{article}
\usepackage{spconf, mathtools, graphicx}
\usepackage{multirow}
\usepackage{slashbox}

	\title{modeling temporal information using discrete fourier transform for recognizing emotions in user-generated videos }
	\author{My Name}
	\name{Haimin Zhang and Min Xu
			}
		
	\address{GBDTC, Faculty of Engineering and IT, University of Technology, Sydney\\
	Haimin.Zhang@student.uts.edu.au, Min.Xu@uts.edu.au}

\begin{document}
\maketitle

\begin{abstract}
With the widespread of user-generated Internet videos, emotion recognition in those videos attracts increasing research efforts. However, most existing works are based on frame-level visual features and/or audio features, which might fail to model the temporal information, \emph{e.g.} characteristics accumulated along time. In order to capture video temporal information, in this paper, we propose to analyse features in frequency domain transformed by discrete Fourier transform (DFT features). Frame-level features are firstly extract by a pre-trained deep convolutional neural network (CNN). Then, time domain features are transferred and interpolated into DFT features. CNN and DFT features are further encoded and fused for emotion classification. By this way, static image features extracted from a pre-trained deep CNN and temporal information represented by DFT features are jointly considered for video emotion recognition. Experimental results demonstrate that combining DFT features can effectively capture temporal information and therefore improve emotion recognition performance. Our approach has achieved a state-of-the-art performance on the largest video emotion dataset (VideoEmotion-8 dataset), improving accuracy from 51.1\% to 55.6\%.
\end{abstract}

\section{Introduction}
\label{sec:intro}

\begin{figure*}[!htd]
\begin{center}
\includegraphics[width=\textwidth-1.2cm]{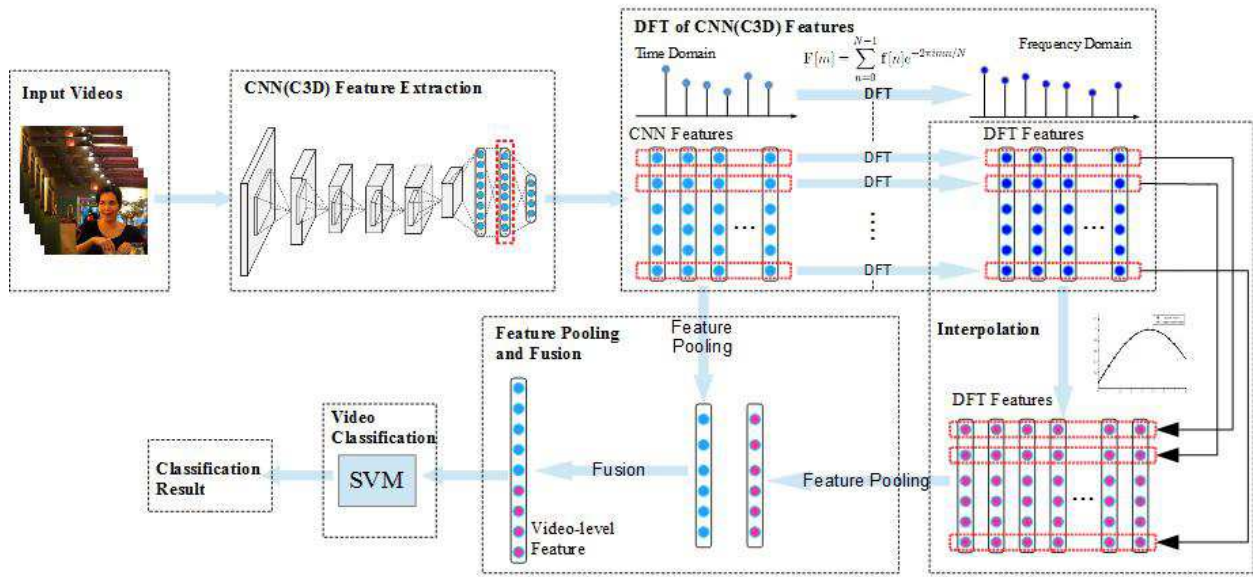}
\caption{An overview of our method for video emotion classification.}
\end{center}
\end{figure*}  	
	With the advances of mobile devices with high quality cameras, people can easily capture, edit and upload video clips to the Internet. It is urgent to develop intelligent video search engines.
Rather than semantic based video searching, people may prefer to pick up their favourite videos according to their emotional preferences since videos may convey different kinds of emotions. Therefore, how to automatically recognise the emotions conveyed in these videos has attracted increasing interests in the research community.
	
While dramatic progress has been achieved in affective image classification \cite{machajdik2010affective, liu2011associating, borth2013sentibank}, few attempts have been made to video emotion recognition. Compared to affective image analysis, video emotion analysis is a complex task. This is mainly because of two reasons: 1) Different from a single image, a video sequence contains multiple frames which group together to reflect a dominant emotion. Within a video sequence, some frames might not convey (or even convey an opposite emotion of) the dominant emotion. 2) As one of the temporal features, characteristics accumulated along time, is difficult to model. Early studies on video emotion recognition mainly focused on movie videos \cite{wang2006affective, xu2013hierarchical}. Most of these studies utilised low-level visual features (\emph{e.g.} colour and brightness) and/or audio features (\emph{e.g.} MFCC and Audio-Six).
In addition to low-level features, Jiang \emph{et. al.} \cite{Jiang2014} introduced so-called attribute features (\emph{e.g.} ObjectBank \cite{li2014object} and Sentibank \cite{borth2013sentibank}) for emotion recognition.

In contrast to hand-crafted features, the last few years has witnessed the success of deep features. Deep features extracted from the activation of a convolutional neural work (CNN) pre-trained on ImageNet \cite{deng2009imagenet} are more discriminative than hand-crafted features \cite{jia2014caffe}. CNN features have achieved state-of-the-art results on many benchmarks \cite{razavian2014cnn} and therefore are widely used in image classification, object detection and attribute detection \cite{zhou2014learning, razavian2014cnn}. Recently, researchers started applying CNN features to video classification \cite{simonyan2014two, wu2015modeling, xu2014discriminative}. 
For example, Xu \emph{et. al.} \cite{xu2014discriminative} proposed a video representation by leveraging frame-level features extracted by CNN with an appropriate feature encoding method.

Most conventional approaches for video classification involve mainly three stages: First, local features or frame-level features are extracted. Then these features are quantised to a fixed length representation using a visual dictionary usually learned by K-means algorithm. Later, video-level representations are obtained by pooling method, such as max-pooling and average-pooling. Lastly, a classifier is trained on the video-level representations to differentiate different classes of video emotions.

However, during the process of feature quantisation and pooling, the temporal information (\emph{i.e.}, information accumulated along time) of videos has not been well treated. Most recently, realising the importance of temporal information for video classification, researchers have started exploring how to take temporal information into account for video analysis.
Lan \emph{et. al.} \cite{lan2014temporal} proposed to model temporal information using temporal division pyramid (TDP) for action recognition. However, the dimension of the video-level features generated by TDP method are very high, which might lead to a problem of classifier overfitting. In \cite{wu2015modeling}, Long Short Term Memory (LSTM) \cite{hochreiter1997long}, which can preserve information for a long time, was adopted to model temporal information for video classification. However, training an LSTM is a time-consuming task.
%

Inspired by DFT, which can transform a discrete signal from time domain to frequency domain, this paper proposes to analysis features in frequency domain to model temporal information for videos. To some extent, signal characteristics along time can be accumulated and represented through sampling in frequency domain. In this work, we integrate CNN features and DFT features for video representation by leveraging feature encoding method.

\section{Overall architecture}
The proposed method consists of five steps as shown in figure 1. The first step is to extract CNN features from video frames. Secondly, considering each feature dimension as a discrete signal over time, we apply DFT to transform the signal to frequency domain.
Thirdly, interpolation method is adopted to generate a fixed length representation for every dimension of DFT features.
Fourthly, locality-constrained linear coding (LLC) method and max-pooling strategy are applied to aggregate CNN features and DFT features. The combination of the aggregated CNN features and DFT features can be regarded as a video-level representation. Finally, with the features generated in the fourth step, an SVM is trained to recognise emotions.
Different from existing methods for temporal information analysis, our method has the following advantages:
\begin{itemize}
\item Compared to LSTM, our method avoids a complex model learning process.
\item Our method captures both spatial information and temporal information with relatively low feature dimensions.
\item Video clips have different length which forms a difficulty for forming a uniformed feature representation. Using DFT is resilient to signal length variation. Moreover, with fast Fourier transform (FFT), computational complexity can be significantly reduced.
\end{itemize}

The rest of this paper is organised as follows. In section 3, the proposed approach is introduced in detail. Experimental results and comparison with previous works are present in section 4. Finally, we conclude this paper in section 5.

\section{The Proposed Algorithm}
In this section, the proposed algorithm is introduced in detail step by step.
\subsection{Feature Extraction}
As shown in \cite{razavian2014cnn}, deep features extracted from a convolutional neural network which is pre-trained on a large image dataset can be used as a powerful feature representation for many visual analysis tasks. In this paper, we leverage a pre-trained deep convolutional neural network \cite{krizhevsky2012imagenet} 
to extract frame-level descriptors for all video clips.
The network consists of five convolution layers and three fully connected layers with a final 1000-way softmax. All input images are resized to 256$\times$256 without considering its original aspect ratio before feeding into the network. Considering features extracted from fully connected layers can capture semantic information for the input image, activation from the last fully connected layer (referred to as fc$_7$ \cite{xu2014discriminative}) are extracted as the frame descriptor. Then \ensuremath{\ell_2} normalization is adopted for all frame-level descriptors. Let $\mathbf{f}$ denote the set of frame-level descriptors of a video clip which has $N$ frames, then $\mathbf{f}$ can be described as
\[
    \mathbf{f}=(\mathbf{f}_1,...,\mathbf{f}_N) =
    \left[
      \begin{array}{cccccc}
        \mathbf{f}_1[1]   & \hdots &\mathbf{f}_i[1] & \hdots & \mathbf{f}_N[1] \\
        \vdots  & \vdots  & \vdots  &  \vdots  & \vdots\\
        \mathbf{f}_1[k]   & \hdots &\mathbf{f}_i[k] & \hdots & \mathbf{f}_N[k] \\
        \vdots  & \vdots  & \vdots  &  \vdots  & \vdots\\
        \mathbf{f}_1[D]   & \hdots &\mathbf{f}_i[D] & \hdots & \mathbf{f}_N[D] \\
      \end{array}
    \right]
    \]
Where $\mathbf{f}_i=(\mathbf{f}_i[1],...,\mathbf{f}_i[D])^{T}$ represents the descriptor of the $i$-th frame with dimension $D$. In this work, $D$ equals 4096, which is the dimension of fc$_7$. The value of $N$ can be different for different video clips.
\subsection{Discrete Fourier Transform of CNN Features}
The aim of discrete Fourier transform (DFT), which is widely used in the field of signal processing, is to transform a discrete signal from time domain to frequency domain. At this step, we present how DFT is applied to CNN features.

As described in 3.1, $\mathbf{f}=(\mathbf{f}_1,...,\mathbf{f}_N)$ represents the set of frame-level CNN features extracted from a video clip with $N$ frames.
Let the $k$-th dimension of $\mathbf{f}$ be denoted $\mathbf{f}[k]=(\mathbf{f}_1[k],\mathbf{f}_2[k],...\mathbf{f}_N[k])$. $\mathbf{f}[k]$ can be considered as a discrete signal which has $N$ sample points with equal sampling time intervals $\Delta t$. We transform $\mathbf{f}[k]$ to frequency domain using the following equation
\begin{equation} \label{eq:DFT}
    \begin{aligned}
     \mathbf{F}_s[k] = \sum_{n=1}^{N}\mathbf{f}_n[k]e^{-2i \pi (n-1)(s-1)/N}, \; s=1,2,...,N.
    \end{aligned}
\end{equation}

Let the result be denoted $\mathbf{F}[k]=(\mathbf{F}_1[k],...,\mathbf{F}_N[k])$, the number of points obtained in frequency domain is as same as that in time domain.
The computed value $\mathbf{F}_s[k] \in \mathbf{F}[k]$ is a complex number and its absolute value represents the amplitude of the $s$-th frequency. In our work, the absolute value of $\mathbf{F}_s[k]$ is used instead of its original complex value. After transforming all $\mathbf{f}[k]$ to frequency domain, where $k=1,...,D$, we get the following feature set
\[
    \mathbf{F}=(\mathbf{F}_1,...,\mathbf{F}_N) =
    \left[
      \begin{array}{cccccc}
        \mathbf{F}_1[1]   & ... &\mathbf{F}_i[1] & ... & \mathbf{F}_N[1] \\
        \vdots  & \vdots  & \vdots  &  \vdots &  \vdots\\
        \mathbf{F}_1[k]   & ... &\mathbf{F}_i[k] & ... & \mathbf{F}_N[k] \\
        \vdots  & \vdots  & \vdots  &  \vdots &  \vdots\\
        \mathbf{F}_1[D]   & ... &\mathbf{F}_i[D] & ... & \mathbf{F}_N[D] \\
      \end{array}
    \right]
    \]
 Where $\mathbf{F}_i=(\mathbf{F}_i[1],...,\mathbf{F}_i[D])^{T}$, termed as a DFT feature in this paper.

\subsection{Interpolation}
As mentioned in section 3.1, the number of sample points $N$ is different due to the various video length.
For two video clips $u$ and $v$ with $N$ and $M$ frames respectively, let $\mathbf{f}^u[k]=(\mathbf{f}^u_1[k],\mathbf{f}^u_2[k],...,\mathbf{f}^u_N[k])$ and $\mathbf{f}^v[k]=(\mathbf{f}^v_1[k],\mathbf{f}^v_2[k],...,\mathbf{f}^v_M[k])$ indicate the $k$-th dimension of CNN features.
We use $\Delta t$ to indicate the sampling time interval which is uniform for all video clips, \emph{i.e.} the sampling rate is $S=1/\Delta t$.

After transforming $\mathbf{f}^u[k]$ and $\mathbf{f}^v[k]$ to frequency domain, we obtain $\mathbf{F}^u[k]=(\mathbf{F}^u_1[k],\mathbf{F}^u_2[k],...,\mathbf{F}^u_M[k])$ and $\mathbf{F}^v[k]=(\mathbf{F}^v_1[k],\mathbf{F}^v_2[k],...,\mathbf{F}^v_M[k])$ respectively.
$\mathbf{F}^v[k]$ and $\mathbf{F}^v[k]$ have the same frequency range from 0 to $S$ with sampling interval $S/N$ and $S/M$ respectively.
From equation \ref{eq:DFT}, we know that the number of points obtained in frequency domain is as same as that in time domain. Therefore signals with more sample points in time domain are more compactly spaced in frequency domain than signals with less sample points.
\begin{figure}[!htbp]
\begin{center}
\includegraphics[width=\textwidth/2-3.0cm]{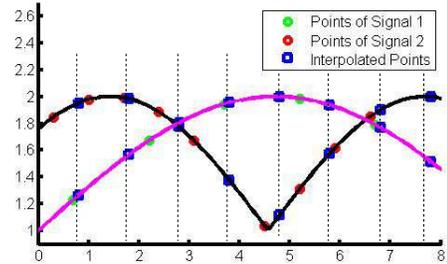}  
\caption{Illustration of applying interpolation to signals with different points in frequency domain.}
\end{center}
\end{figure}

Like image resizing, we use cubic interpolation method \cite{keys1981cubic} to generate a fixed length representation $\mathbf{F}^u[k]=(\mathbf{F}^u_1[k],\mathbf{F}^u_2[k],...,\mathbf{F}^u_L[k])$ and $\mathbf{F}^v[k]=(\mathbf{F}^v_1[k],\mathbf{F}^v_2[k],...,\mathbf{F}^v_L[k])$. By this way, different video clips have the same frequency sample interval from the frequency range from 0 to $S$, as shown in Figure 2.

\subsection{Feature Encoding and Fusion}
In order to generate a uniform video-level representation, we need to aggregate the obtained CNN features and DFT features separately. Firstly, locality-constrained linear coding (LLC) algorithm \cite{wang2010locality} is adopted for feature embedding. LLC is an extension of the sparse coding method. It utilises locality constraints, which selects k-nearest codewords from a dictionary learned by K-means algorithm, and generates a sparse representation for the input vector.
After that, we choose to use max-pooling strategy to aggregate the LLC-based features.

We then adopt late feature fusion. The linear combination of the aggregated LLC-based CNN features and DFT features lead to the final video-level representation.
\subsection{Emotion Classification}
After obtaining all video-level features, an SVM is trained by optimising the following equation \cite{fan2008liblinear} for emotion classification.
\begin{equation} \label{eq:SVM}
\mathrm{arg}\,\mathrm{min_{\textbf{w},b}}\; \frac{1}{2} \|\textbf{w}\|^2+C\sum_{i=1}^{T}max(1-y_i(\textbf{w}^T \textbf{x}_i) + b),0)
\end{equation}
Where $\textbf{x}_i$ and $y_i$ represent video-level feature and its corresponding label, $C$ and $T$ indicate penalty parameter and the number of training features respectively.

\section{Experiments}
\subsection{Experimental Setup}
Experiments were conducted on the VideoEmotion-8 dataset \cite{Jiang2014}, which contains 1,101 user-generated videos labelled with 8 basic human emotion categories. Each category is with a minimum number of 100 videos per class and an average duration of 107 seconds. Currently, this dataset is the largest dataset for recognising emotions in user-generated videos.

Similar as \cite{Jiang2014}, we ran test for 10 times. For each time, we randomly selected $2/3$ data from each class for training and the rest for testing. The average accuracy was further calculated to evaluate the classification performance.

For computation efficiency, we sampled a frame every 8 frames.
Frame-level descriptors were extracted using the Caffe toolkit \cite{jia2014caffe}.  We selected 50 videos randomly from each emotion category for fine-tuning. The CNN model was first pre-trained on ImageNet \cite{deng2009imagenet} and then fine-tuned by using the video frames. All frame-level descriptors were further \ensuremath{\ell_2} normalised.
Fast Fourier transform (FFT) \cite{frigo1998fftw} was adopted for computing DFT. At the interpolation step, $L$ was set to 500 experimentally.
For LLC coding, we trained a vocabulary with 1024 codewords for CNN features and DFT features separately. The aggregated LLC-based CNN features and DFT features were normalised to 3$/$5 and 2$/$5 respectively using \ensuremath{\ell_2} normalisation.

We applied the LibLinear toolbox \cite{fan2008liblinear} for SVM classification. The penalty parameter $C$ was set to 1 experimentally.

\subsection{Experimental Results and Discussion}
In our experiments, we intended to evaluate: (1) the performance of CNN features; (2) the performance of DFT features; and (3) the overall performance of combined CNN and DFT features. Moreover, to prove the efficiency of our proposed method, we compared our results with the most recent two works \cite{Jiang2014} and \cite{pang2015deep}. Experimental results are shown in Table 1, from which we can find:
\begin{table}[!tbp]
\centering  
\begin{tabular}{|c|c|c|c|c|c|}  
\hline
Category &Jiang&Pang &CNN &DFT &CNN+DFT\\
&\cite{Jiang2014} &\cite{pang2015deep} && &\\\hline
Anger        &53.0  &48.5  &58.6  &53.7  &62.1\\         
Anticipation &7.6   &0   &41.6  &32.7  &42.4\\        
Disgust      &44.6  &53.8  &40.1  &44.3  &50.3\\
Fear         &47.3  &52.7  &50.7  &44.0  &51.8\\
Joy          &48.3  &54.2  &45.1  &46.7  &53.7\\
Sadness      &20.0  &32.4  &60.0  &43.1  &60.9\\
Surprise     &76.9  &78.7  &74.4  &66.0  &76.2\\
Trust        &28.5  &43.8  &47.8  &24.6  &47.3\\
\hline
Overall      &46.1  &51.1  &\textbf{52.3}  &\textbf{44.4}  &\textbf{55.6}\\
\hline
\end{tabular}
\caption{Prediction accuracy (\%) of each emotion category using CNN, DFT and the concatenation of the two features, and comparison with previous works.}
\end{table}
\begin{enumerate}
  \item[(1)] In comparison with \cite{Jiang2014} and \cite{pang2015deep}, the accuracy of our method using CNN features improves 6.2\% and 1.2\% respectively. In \cite{Jiang2014} and \cite{pang2015deep}, the authors used low-level visual features, audio features and attribute features, whereas CNN features were applied in our work. Experimental results demonstrate that the performance of CNN features may be superior than hand-crafted features.
  \item[(2)] Using DFT features only can not improve the classification performance.
  \item[(3)] Combining CNN and DFT features, our method outperforms \cite{Jiang2014} 9.5\% and \cite{pang2015deep} 4.5\%. Overall, incorporating DFT features, the classification performance can be improved 3.3\% compared with applying CNN feature only. The results prove the efficiency of the DFT features. The major reason might be that we modeled temporal information by analysing DFT features. By combining temporal information, the recognition accuracy can be improved to some extent. Another possible reason is that LLC was used for feature encoding, which might had better performance than bag of words model that was used in \cite{Jiang2014}. To the best of our knowledge, this is the state-of-the-art performance on VideoEmotion-8 dataset.
\end{enumerate}
\section{Conclusions}
In this paper, we have proposed to apply CNN and DFT features to model spatial and temporal information for video emotion analysis. Compared to hand-crafted features, CNN features can improve the classification accuracy significantly. Capturing temporal information, DFT features have been further proved to be efficient and effective. The combination of CNN and DFT features achieves the state-of-the-art performance on VideoEmotion-8 dataset.
\bibliographystyle{IEEEbib}
\bibliography{refs}
\end{document}